\documentclass{article}
\pdfoutput=1

\usepackage{arxiv}

\usepackage[utf8]{inputenc} % allow utf-8 input
\usepackage[T1]{fontenc}    % use 8-bit T1 fonts
\usepackage{hyperref}       % hyperlinks
\usepackage{url}            % simple URL typesetting
\usepackage{booktabs}       % professional-quality tables
\usepackage{amsfonts}       % blackboard math symbols
\usepackage{nicefrac}       % compact symbols for 1/2, etc.
\usepackage{microtype}      % microtypography
\usepackage{lipsum}		% Can be removed after putting your text content
\usepackage{graphicx}
\usepackage{natbib}
\usepackage{doi}
\usepackage{soul}
\usepackage{algorithm}
\usepackage{algpseudocode}

\usepackage{times}
\usepackage{latexsym}
\usepackage{multirow}
\usepackage{color, colortbl}
\usepackage[T1]{fontenc}
\usepackage[utf8]{inputenc}
\usepackage{microtype}
\usepackage{inconsolata}

\title{Metaphorical Paraphrase Generation: Feeding Metaphorical Language Models with Literal Texts}

\author{ {\href{https://orcid.org/0000-0002-0822-0345}{\includegraphics[scale=0.06]{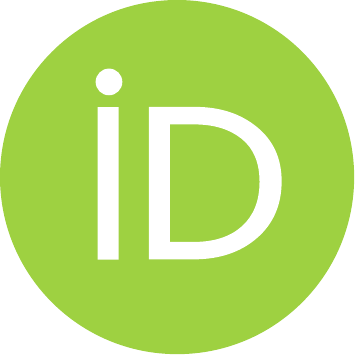}\hspace{1mm}Giorgio Ottolina}} \\%\thanks{Use footnote for providing further
	Department of Informatics,\\
	Systems and Communication\\
	University of Milano-Bicocca\\
	Milan, Italy \\
	\texttt{g.ottolina1@campus.unimib.it} \\
	\And
	\href{https://orcid.org/0000-0001-9188-7425}{\includegraphics[scale=0.06]{orcid.pdf}\hspace{1mm}John Pavlopoulos} \\
	Department of Computer and Systems Sciences\\
	Stockholm University\\
	Stockholm, Sweden \\
	\texttt{ioannis@dsv.su.se} \\
}

\renewcommand{\shorttitle}{\textit{arXiv} Template}

\hypersetup{
pdftitle={A template for the arxiv style},
pdfsubject={q-bio.NC, q-bio.QM},
pdfauthor={David S.~Hippocampus, Elias D.~Striatum},
pdfkeywords={First keyword, Second keyword, More},
}

\begin{document}
\maketitle

\begin{abstract}
This study presents a new approach to metaphorical paraphrase generation by masking literal tokens of literal sentences and unmasking them with metaphorical language models. Unlike similar studies, the proposed algorithm does not only focus on verbs but also on nouns and adjectives. Despite the fact that the transfer rate for the former is the highest (56\%), the transfer of the latter is feasible (24\% and 31\%). Human evaluation showed that our system-generated metaphors are considered more creative and metaphorical than human-generated ones while when using our transferred metaphors for data augmentation improves the state of the art in metaphorical sentence classification by 3\% in F1.
\end{abstract}

\section{Introduction}

Figurative language is ambiguous and often contains mapping of concepts from one domain to another. In ``\textit{The wheels of Stalin's regime were well-oiled and already turning}'', for example, a political system is viewed through the lens of conceptual metaphor theory \cite{lakoff} as a mechanism which can function, break, and have wheels. Due to its challenging nature, even Transformers struggle to model figurative language \cite{Chakrabarty2022}. This, however, is not only hindering the progress of computational metaphor detection, but also that of Natural Processing tasks, such as sentiment analysis \cite{Liu2020}, or other ones, such as cybersecurity \cite{hilton2022metaphor}. %Pre-training on and transfer learning from large corpora could potentially alleviate this problem, but such resources are not currently available. With this work we propose an alternative 

%In order to better understand metaphors and their complexity, as well as the challenges that they can bring to natural language processing tasks, it is important to look at practical examples and at the related core literature studies. Consider, for instance, the following metaphorical sentence: \textit{The wheels of Stalin's regime were well-oiled and already turning}, where a political system is viewed in terms of a mechanism which can function, break, have wheels, etc. This association allows us to transfer knowledge from the domain of \textit{mechanisms} to that of \textit{political systems}. Therefore, political systems are thought about in terms of mechanisms, and discussed through the mechanism terminology, leading to multiple metaphorical expressions. This particular view of metaphors is known as Conceptual Metaphor Theory, and it was first introduced by \citet{lakoff} in 1980. There are different types of metaphors, such as the \textbf{is-a} type (e.g., \textit{That lawyer is a shark}), the  \textbf{of} type (e.g., \textit{Child of evil}), or \textbf{verb}-based (e.g., \textit{He cut me off, yet still I carried his name}).

Computational approaches related to metaphors that can be found in literature mostly focus on detection and generation. Metaphor detection comprises the \textit{identification} of metaphor-related words in the text \cite{fass,Birke,shutova4,Steen} and \textit{interpretation}, which is mostly based on paraphrasing \cite{advances}. Metaphor \textit{generation} concerns the task of creating novel metaphorical sentences, for example by taking literal ones and transforming them in a way that makes them acquire a figurative meaning, which can be useful to poetry generation \cite{van-de-cruys-2020}.
No study in literature, to the best of the authors' knowledge, however, has tried to simultaneously address metaphor detection and generation in the same setting. 
Furthermore, all existing metaphor generators \cite{MERMAID,Yu,advances,Brooks,Stowe2021} depend on custom and external systems that go beyond standard fine-tuning procedures.

\begin{center}
    \begin{figure}[pt]
        \centering
        \includegraphics[width=.48\textwidth]{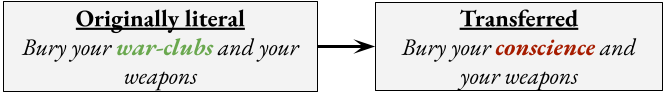}
        \caption{Literal sentence (on the left) transferred to a metaphorical one (right) with the automated substitution of a single term.} 
        \label{fig:framework_highlevel}
    \end{figure}
\end{center}

With this work we present a new metaphorical language generation perspective by transferring literal to metaphorical texts. The transfer concerns the replacement of tokens of different parts of speech, not only the common verb type \cite{Stowe2021,MERMAID,Yu}, but also nouns and adjectives (examples shown in Fig.~\ref{fig:framework_highlevel}). Human evaluation showed that a randomly selected sample of our system-generated metaphors were \textit{more creative and metaphorical} than respective human-generated ones. Our method is open-source and can be exploited to generate an infinite number of new metaphors, assisting tasks such as metaphor detection through data augmentation. Experimenting with this hypothesis, we show that when we used our system-generated metaphors to augment the training data, the performance of a state of the art metaphor detector improves by 3\%.

\section{Related Work}\label{sec:related_work}
Metaphor identification is often treated as a sequence labelling task, creating an output that consists of a sequence of labels (metaphorical or not) for a sentence or a sequence of input words~\cite{bizzoni-ghanimifard-2018-bigrams,chen-etal-2020-go,dankers-etal-2020-neighbourly,gao1,gong-etal-2020-illinimet,mao-etal-2019-end,mykowiecka-etal-2018-detecting,pramanick-etal-2018-lstm,su-etal-2020-deepmet,wu-etal-2018-neural}.
%The first sequence labelling approaches usually represented an input sentence as a concatenation of pre-trained word embeddings and generated a context-specific sentence embedding exploiting bidirectional long short-term memory, or \textit{BiLSTM}~\cite{dankers-etal-2020-neighbourly,gao1,mykowiecka-etal-2018-detecting,pramanick-etal-2018-lstm,bizzoni-ghanimifard-2018-bigrams}.
%BiLSTM systems take advantage of both contextualised and pre-trained embeddings in the classification layer. 
\textit{Di-LSTM} Contrast~\cite{swarnkar-singh-2018-di} encoded the left and right side context of a target word through forward and backward LSTMs. 
%The classification is based on a concatenation of the target word representation and its difference with the encoded context~\cite{advances}. 
\citet{mao-etal-2019-end} combined GloVe~\cite{glove} and BiLSTM hidden states for sequence labelling. 
%Static embeddings like GloVe~\cite{glove} do not change with the context once been learned. 
Despite their efficiency, the static nature of embeddings such as GloVe makes it difficult to cope with polysemy, which is crucial when dealing with metaphors since the meaning of a polysemous word depends on its context~\cite{din_emb}.
%To deal with the problem of polysemy, a number of approaches have been recently proposed to learn the representation of words among their context. For example, in the following two sentences: “Apple sells phones” and “I eat an apple”, dynamic embeddings~\cite{din_emb} will represent “apple” differently according to the context, while static embeddings can not distinguish the semantic difference between the two references of “apple”. Dynamic embeddings extracted from pre-trained language models~\cite{bert,contextual,contextual2,radford} have demonstrated dramatic superiority over their static predecessors in various NLP tasks, and also in metaphor detection and generation approaches.
Fine-tuning pre-trained contextual language models, however, do not suffer from this problem~\cite{chen-etal-2020-go,dankers-etal-2020-neighbourly,gong-etal-2020-illinimet}. %\citet{dankers-etal-2020-neighbourly}, for example, fine-tuned a BERT model, which gets a discourse fragment as input. 
%Hierarchical attention computes both token and sentence level attention~\cite{attention} after the encoded layers, leading to better results compared to those obtained by applying general attention to all tokens.

\begin{table}[ht]
    \centering\small
     \caption{Metaphor generation studies that comprise masked language modeling (MLM), metaphor reconstruction (MR), and/or recognition of the metaphor's location within the text (MLR).}
    \begin{tabular}{|c|c|c|c|c|c|}
    \hline
        \textbf{}  & \textbf{MLM} & \textbf{MR} & \textbf{MLR} \\ \hline
        \textbf{\citet{MERMAID}} & \textbf{X} & \textbf{X} &  \\ \hline
        \textbf{\citet{Yu}} & & \textbf{X} & \textbf{X} \\ \hline
        \textbf{\citet{Brooks}} & & \textbf{X} & \\ \hline
        \textbf{\citet{Stowe2021}} & & \textbf{X} & \\ \hline\hline
        \textbf{\textbf{Ours}} & \textbf{X} & \textbf{X} & \textbf{X}\\\hline
    \end{tabular}
\label{tab:related_work}
\end{table}

Metaphor generation is usually based on obtaining novel figurative sentences either by replacing verbs contained in literal phrases \cite{MERMAID,Yu,Stowe2021}, or by exploiting syntactic patterns that discriminate between creative metaphorical expressions and non-metaphorical ones \cite{Brooks}. Information regarding recent advances in metaphor detection, processing and generation, can be found in \cite{advances} while Table~\ref{tab:related_work} presents the main studies. \citet{MERMAID} fine-tuned BART~\cite{Lewis2020} on a parallel corpus of metaphorical and literal sentences, which they created by replacing relevant verbs from literal expressions and by applying masked language modeling (MLM) \cite{Singh} on metaphorical sentences from the Gutenberg Poetry corpus~\cite{gutenberg}. We also address the lack of resources by employing MLM and reconstruction, but our approach does not need a parallel corpus while it can recognise the metaphor's location and hence it is not limited to verbs.

\citet{Yu} employed a neural approach to extract the metaphorical verbs from the sentences along with their metaphorical senses in an unsupervised way. Then, the same neural approach is exploited to train a neural language model from a Wikipedia corpus. The novel metaphors are obtained by conveying the assigned metaphorical senses through a decoding algorithm. \citet{Stowe2021} obtained new metaphorical sentences by replacing relevant verbs in literal expressions and encoding conceptual mappings (FrameNetbased embeddings - \textit{CM-LEX}, and a custom seq-to-seq model - \textit{CM-BART}) between cognitive domains. \citet{Brooks} trained an unsupervised LSTM model and used an inherent inference engine to create new metaphors. The novelty of these new metaphors is ensured by checking that none of the generated sentences match the training data, and that the identified syntactic patterns of metaphors were not present in the non-metaphorical data. Besides focusing on more than verbs and disregarding language-specific syntactic patterns, our approach does not depend on external or custom systems.

\pagebreak

\section{Literal to Metaphor Transferring} \label{sec:framework}

The proposed literal to metaphor generation approach is described with Algorithm~\ref{alg:mdg}. 

\begin{figure*}[ht]
    \centering
    \includegraphics[width=.8\textwidth]{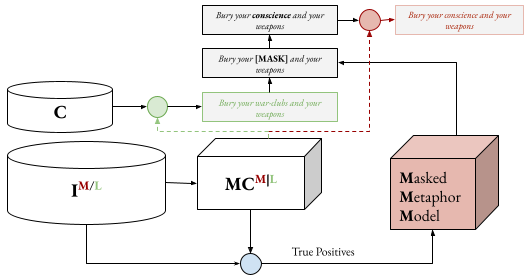}
    \caption{Depiction of the proposed metaphor generation process. The input consists of a set of unlabelled sentences ($C$) and a set of sentences that are labelled as metaphorical or literal ($I^{M/L}$), following Algorithm~\ref{alg:mdg}.} 
    \label{fig:framework}
\end{figure*}

The concept of reconstructing metaphors is not new, and it was first used in~\cite{sullivan2007grammar}.
Algorithm~\ref{alg:mdg}'s input consists of a corpus $C$ of unlabelled texts, a dataset $I^{M|L}$ of texts labelled as literal or metaphorical, and a classification threshold $h$.
We refer to “M|L” as the sentence’s label being either metaphorical or literal. True positives based on $MC^{M/L}$ are used to fine-tune a masked metaphor model (MMM).

\noindent\textbf{Metaphorical text classification} is a binary classification task (Algorithm~\ref{alg:mdg}, line 2), where a sentence is considered metaphorical if the returned score is higher than a threshold (e.g., in line 7) and literal otherwise (e.g., line 4). A classifier can be trained on a dataset with a binary label per text, such as $I^{M|L}$.

\noindent\textbf{Masked metaphor modeling} concerns the restoration of masked metaphors (line 1). This is a task similar to masked language modeling but the masked tokens are tagged metaphors. Fine-tuning a pre-trained model requires a limited number of texts classified as metaphorical and with the metaphor tagged within the text. The algorithm assumes that $I^{M|L}$ comprises such tags. 

\noindent\textbf{Generation of new metaphors} requires sentences from a corpus $C$ which $MC^{M/L}$ has classified as literal (line 4). A random token is masked and the masked metaphor model replaces the masked token, effectively trying to create a metaphorical text. There is no guarantee that the masked token of the literal sentence will be replaceable by one that would turn the text into metaphorical. Hence, we employ again $MC^{M/L}$ now to keep only truly transferred texts (line 7). 

\noindent\textbf{The overall process} is also depicted in Figure~\ref{fig:framework}. A classifier, named $MC^{M/L}$, is trained on $I^{M/L}$. True positives based on $MC^{M/L}$ are used to fine-tune a masked metaphor model (MMM). MMM is then used to replace masked tokens of sentences from $C$, originally classified by $MC^{M/L}$ as literal. Reconstructed sentences, which are also classified as metaphorical by $MC^{M/L}$, are finally returned by the algorithm.

\begin{algorithm}
\caption{Literal To Metaphor}\label{alg:mdg}
\begin{algorithmic}[1]
\Require: A corpus $C$ of unlabelled texts, a dataset $I^{M|L}$ of texts labelled as literal or metaphorical, and a classification threshold $h$.
\Ensure A set of metaphorical sentences.
\State $MR \gets finetune(I^{M|L}, \textit{reconstruction})$
\State $MC^{M|L} \gets finetune(I^{M|L}, \textit{classification})$

\While{$N \neq 0$}

\If{$MC^{M|L}(t \in C)<h$}

\State $t^{mask} \gets mask(t)$ \Comment{Random mask}
\State $t^{met} \gets MR(t^{mask})$ \Comment{Reconstruct}

\If{$MC^{M|L}(t^{met})>h$}

\State $yield\ t^{met}$
\EndIf
\EndIf
\EndWhile
\end{algorithmic}
\end{algorithm}

\section{Empirical analysis} \label{sec:empirical_evaluation}
We undertook an empirical analysis on the most common datasets employed for metaphor-related tasks and experimented with several baselines. 

\subsection{Datasets}\label{ssec:data}
Table~\ref{tab:dataset} presents the three datasets we experimented on: MOH-X, TroFi, and TroFi-X. 

\noindent\textbf{MOH-X}~\cite{MOH-X} is derived from the subset of the MOH dataset that was used by \citet{shutova3}.
\citet{MOH-X} annotated different verbs for metaphoricity.
They extracted verbs that had between three and ten senses in WordNet~\cite{WordNet} along with their glosses.
The verbs were annotated for metaphoricity with the help of crowd-sourcing. Ten annotators were recruited to assess each sentence, and only those verbs that were annotated as positive for metaphoricity by at least 70\% of the annotators were selected in the end.
The final dataset consisted of 647 verb-noun pairs: 316 metaphorical, and 331 literal.

\noindent\textbf{TroFi} contains feature lists consisting of the stemmed nouns and verbs in a sentence, with target or seed words. It is named after TroFi (Trope Finder), a nearly unsupervised clustering method for separating literal and non-literal usages of verbs~\cite{Birke}. For example, given the target verb \textit{pour}, TroFi is able to cluster the sentence \textit{Custom demands that cognac be poured from a freshly opened bottle} as literal, and the sentence \textit{Salsavand rap music pour out of the windows} as nonliteral.
The target set is built using the `88-`89 Wall Street Journal Corpus \footnote{https://catalog.ldc.upenn.edu/LDC2000T43}  tagged using the \citet{Ratnaparkhi} tagger  and the \citet{Joshi} SuperTagger . The final dataset consisted of 3,737 sentences.

\noindent\textbf{TroFi-X} is an alternative version of \textbf{TroFi}. It contains 1,444 sentences annotated not only with metaphorical verbs, but also with metaphorical nouns, pronouns and adjectives.

\begin{table}[]
    \centering
     \caption{Statistics of all the datasets employed in this work. All datasets comprise English sentences. Size is measured in sentences and POS shows the part(s) of speech of the metaphors.}
    \begin{tabular}{|c|c|c|c|c|p{4cm}|}
    \hline
        \textbf{Name} & \textbf{Size}  & \textbf{POS} \\ \hline
        \textsc{moh-x} & 646 & Noun/Verb  \\ \hline
        \textsc{TroFi} & 3,737 & Verb \\ \hline
        \textsc{TroFi-X} &1,444 & Noun/Verb/Adjective \\
     \hline
    \end{tabular}
\label{tab:dataset}
\end{table}

\subsection{Evaluation measures}
For the classification task, we employed Accuracy (i.e. the fraction of instances that were correctly classified), Precision (i.e., the number of instances that were correctly predicted as metaphorical to the number of instances that were predicted as metaphorical), Recall (i.e., the number of instances correctly predicted as metaphorical to the number of instances that should have been predicted as metaphorical) and F1 (i.e., the harmonic mean of Precision and Recall).
For the reconstruction task, we employed Accuracy (i.e., the ratio of sentences that are correctly reconstructed/generated).

\subsection{Methods}\label{ssec:methods}

\begin{table*}[]
    \centering\small
    \begin{tabular}{c|c|c|c|c||c|c|c|c||c|c|c|c|}
         & \multicolumn{4}{c}{\sc moh-x} & \multicolumn{4}{c}{\sc trofi} & \multicolumn{4}{c}{\sc trofi-x} \\
         &P&R&F1&Ac&P&R&F1&Ac&P&R&F1&Ac\\\hline
         \sc bert&73.91&80.55&79.45&76.92&39.82&91.21&55.44&41.98&66.50&63.08&65.58&69.17\\
         \sc bert(ft)&88.11&\bf94.07&\bf91.86&\bf91.74&72.00&64.24&67.77&72.08&70.18&69.85&70.53&74.91\\
         \sc xlm-r&63.53&72.22&69.33&64.62&58.58&59.46&64.94&74.60&58.98&60.00&63.93&69.66\\
         \sc xlm-r(ft)&\bf88.57&86.11&87.32&86.15&\bf95.86&\bf93.92&\bf94.88&\bf95.99&\bf92.42&\bf93.85&\bf93.13&\bf93.79\\\hline
         \sc nb&73.00&12.33&17.64&65.22&72.73&11.27&19.51&64.71&77.11&80.23&78.54&78.17\\
         \sc rf&61.90&81.25&70.27&65.62&67.50&38.03&48.65&69.52&71.22&87.31&78.66&79.43\\
         \sc knn&65.00&81.25&72.22&68.75&63.29&70.42&66.67&73.26&78.58&81.26&80.09&80.19\\
         \sc svm&66.67&37.50&48.00&59.38&69.11&76.26&72.43&74.32&77.15&65.38&70.62&77.39\\
         \sc lr&87.50&87.50&87.50&87.50&84.51&84.51&84.51&88.24&77.00&81.00&79.00&79.00\\
         \sc mlp&67.00&71.00&69.00&69.00&55.00&44.00&49.00&58.00&71.00&63.00&67.00&73.00\\
    \end{tabular}
    \caption{All tested Classifiers and their respective results for the three metaphorical data sources. \textit{P, R, F1 and Ac} stand respectively for Precision, Recall, F1 Score and Accuracy. Along with \textit{BERT} and \textit{XLM-R} (base and fine-tuned), we have, in order: \textit{NB} - Naive Bayes; \textit{RF} - Random Forest; \textit{KNN} - K-nearest Neighbours; \textit{SVM} - Support Vector Machine; \textit{LR} - Logistic Regression; \textit{MLP} - Multi-Layer Perceptron Neural Network.}
    \label{tab:classification}
\end{table*}

For metaphorical sentence classification, we employed Naive Bayes~\cite{bayes}, Random Forests~\cite{forests}, KNN~\cite{knn}, SVM~\cite{svm}, Logistic Regression~\cite{logreg}, MLP~\cite{mlp}, BERT~\cite{bert}, and XLM-R~\cite{xlm-r}.

To build our metaphor reconstruction model, we used sentences where the (location of the) metaphor was known and which our text classifier $MC^{M|L}$ scored as metaphorical (true positives in Fig.\ref{fig:framework}). We used these sentences to fine-tune BART, T5, BERT and XLM-R. After masking the known metaphor in each sentence, the former two reconstructed and the latter performed masked (metaphorical) language modeling in order to learn to generate the missing metaphorical part.

\subsection{Experimental Results}
Table~\ref{tab:classification} provides the results of the metaphorical sentence classifiers (see Section~\ref{ssec:methods}) on the three metaphorical data sources (see Section~\ref{ssec:data}). XLM-R (fine-tuned) has the best Precision in all datasets. BERT (fine-tuned) achieves the best Recall on MOH-X, leading also to the best Accuracy and F1. Overall, BERT and XLM-R (fine-tuned) yield the best results.
Naive Bayes, Random Forests, KNN, SVM and MLP performed much lower. However, it is worth noting that Logistic Regression, despite its simple nature, performed surprisingly well.

\begin{table*}[]
    \centering\small
    \begin{tabular}{c|c|c||c|c||c|c||c|c|c||c|c|c}
         & \multicolumn{4}{c}{\sc moh-x} & \multicolumn{2}{c}{\sc trofi} & \multicolumn{6}{c}{\sc trofi-x} \\
         &\multicolumn{2}{c}{\sc bert} & \multicolumn{2}{c}{\sc xlm-r} & \multicolumn{1}{c}{\sc bert} & \multicolumn{1}{c}{\sc xlm-r} & \multicolumn{3}{c}{\sc bert} & \multicolumn{3}{c}{\sc xlm-r}\\\hline
         &N&V&N&V&V&V&T1&T2&V&T1&T2&V\\\hline
         T5 &\bf80.65&\bf93.55&83.87&96.77&\bf95.38&\bf96.92&77.14&82.86&88.57&68.57&85.71&\bf97.14  \\
         BART &64.67&90.32&\bf84.62&\bf96.92&\bf95.38&95.38&64.62&\bf83.08&\bf93.85&69.73&\bf87.69&95.38\\
         MMM(ft) &71.43&48.57&77.42&45.16&74.29&83.87&\bf85.71&77.78&58.33&\bf94.44&86.11&66.67\\
    \end{tabular}
    \caption{Accuracy of T5, BART, and two MMMs (BERT, XLM-R) used to reconstruct metaphorical tokens on three datasets. Only sentences correctly classified as metaphorical (by BERT and XLM-R sentence classifiers) are used. Noun (N) and verb (V) accuracy scores indicate the percentage of correctly reconstructed metaphorical nouns and verbs, respectively. TroFi-X sentences comprise three metaphorical tokens each. The first two, \textit{T1} and \textit{T2}, can be of any part-of-speech while \textit{V} is always a verb. The best one per column is shown in bold.}
    \label{tab:reconstruction}
\end{table*}

Table~\ref{tab:reconstruction} presents the accuracy in metaphor reconstruction on the metaphorical sentences that have been correctly classified as metaphorical (the green box in the middle, in Fig.~\ref{fig:framework}) by the best-performing fine-tuned BERT and XLM-R (see Table~\ref{tab:classification}). We employed T5 and BART, as well as two masked language models, BERT and XLM-R~\cite{bert-mlm,xlm-mlm}, which have been fine-tuned by masking (known) metaphorical tokens of the metaphorical sentences. We refer to this process as Masked Metaphor Modeling (MMM; the red box on the right of Fig.~\ref{fig:framework}).
MMM with BERT was applied only on sentences correctly classified as metaphorical by BERT while MMM with XLM-R was applied on sentences correctly classified by XLM-R. T5 and BART were applied on both and results are shown in respective columns (see Table~\ref{tab:reconstruction}).
In MOH-X, the accuracy scores for \textit{nouns} and \textit{verbs} show the percentage of correctly reconstructed metaphorical tokens (respectively nouns or verbs) inside the sentences, by the different reconstruction models. TroFi sentences comprise only verb metaphors while TroFi-X sentences comprise three metaphorical tokens each; the first two, \textit{T1} and \textit{T2}, can be any part-of-speech tokens, while \textit{V} can only be verb metaphors.\footnote{The following sentence taken from TroFi-X is given as an example: \textit{Beyond that, conditions on board were so vile that `` the sailor was at greater \textbf{risk eating} his \textbf{meals} aboard than fighting. ''}. Here, \textbf{risk} is ``token 1'' (in this case, it is a noun), \textbf{meals} is ``token 2'' (in this case, also a noun), and \textbf{eating} is the verb token of the sentence (one of the three metaphorical tokens in each TroFi-X sentence is always a verb).}
    
MMM with XLM-R is consistently better than that with BERT. This is also true for MOH-X, where BERT outperforms XLM-R for metaphorical sentence classification (see Table~\ref{tab:classification}), which means that XLM-R is better in reconstruction. BART and T5 are also overall better when metaphorical sentence classification has been performed with XLM-R. 
When focusing on results obtained using XLM-R as the metaphorical sentence classifier, nouns are more accurately reconstructed by BART on MOH-X and TroFi-X (for T2). T5, which achieves a high accuracy in all datasets as far as verb reconstruction is concerned, is better than BART in TroFi and TroFi-X and only slightly worse in MOH-X.
When comparing MMM with T5 and BART, the latter two seem to work better across MOH-X and TroFi sentences. MMM models, however, perform better on the first tokens (\textit{T1}) of TroFi-X sentences.

\section{Building synthetic metaphors} \label{sec:discussion}
The proposed algorithm required two training steps. First, a text classifier learns to  classify metaphorical from literal sentences. Second, metaphorical sentences,\footnote{The location of the metaphor is assumed known.} which were correctly classified by our classifier as metaphorical, were passed to a reconstruction step, where metaphorical tokens were masked and then recovered through extraction (T5, BART) and masked language modeling (BERT, XLM-R). The algorithm then receives a corpus of literal sentences, masks a random token (Algorithm~\ref{alg:mdg}, line 5), and replaces it with a metaphorical one, inferring new metaphorical sentences created from originally literal ones. Our experimentation with this inference part is described next. 

As literal sentences we used: 2,000 sentences scraped from Wikipedia, related respectively to music (1,000 sentences) and technology (1,000 sentences) topics; 1,000 sentences scraped from the Gutenberg Poetry Corpus~\cite{gutenberg}, which comprises 3,085,117 lines of poetry extracted from hundreds of books.
We applied our fine-tuned XLM-R classifier (Table~\ref{tab:classification}) on these sentences, and the ones classified as literal (Algorithm~\ref{alg:mdg}, line 4) were fed into our XLM-R-based MMM (Table~\ref{tab:reconstruction}). Filtering out incorrectly transferred sentences (line 7), the algorithm yields new metaphorical sentences (line 8). 

Table~\ref{tab:wikipedia} presents the ratio of originally literal sentences that have been (automatically) classified as metaphorical, after replacing a randomly selected (literal) noun, verb or adjective with a metaphorical token. Higher ratios are preferred, because they indicate a successful transfer based on the employed classifier. When the token to be replaced by the MMM was a verb, more than 50\% of the literal sentences from the Gutenberg Poetry Corpus and 43\% of the Wikpedia sentences related to music were turned into metaphorical ones. When the token was an adjective, the ratios dropped to 27\% and 31\% respectively. The lowest ratios were obtained for nouns, where 24\% of the Gutenberg and 22\% of the Wikipedia (related to music) sentences were transferred. Wikipedia sentences related to technology had the lowest ratios of all, achieving 29\% for verbs but 8\% for nouns and 7\% for adjectives.\footnote{We note that in principle, any number of new metaphorical sentences can be generated given any positive ratio. For example, the proposed algorithm can be applied on more literal sentences to counter-balance a low ratio.}

\begin{table}
  \centering\small
  \caption{Ratio of literal sentences that were classified as metaphorical, after applying MMM on a verb, noun, or adjective per sentence. XLM-R used in both tasks.}
    \begin{tabular}{ c c c c }
    \hline
    & Nouns & Verbs & Adj. \\
    \hline
    
    Wikipedia - Music &0.22 &0.43 &\textbf{0.31}  \\
    Wikipedia - technology &0.08 &0.29 &0.07 \\
    Gutenberg Poetry Corpus &\textbf{0.24} &\textbf{0.56} &0.27\\
    
    \hline
  \end{tabular}
  \label{tab:wikipedia}
\end{table}

%The last important step was the one regarding the new metaphorical sentences' evaluation. 
Following the work of ~\citet{MERMAID}, we performed human evaluation of the newly constructed metaphorical sentences, by comparing them against human-generated ones. Two hundred metaphorical sentences were selected, 100 that were built with our algorithm, starting from sentences that originally came from both Wikipedia and Gutenberg Poetry Corpus data sources, and 100 from our employed metaphorical datasets. Two graduates of English literature were then asked to evaluate each sentence. We asked the annotators to assess \textit{How metaphoric are the generated utterances} and named this dimension metaphoricity. Tokens that were supposedly being used in a figurative way inside the sentences were shown in bold and sentences were shuffled. For each sentence, then, four different dimensions were evaluated, vis. fluency, meaning, creativity, and metaphoricity.
\begin{enumerate}
    \item \textit{Fluency (Flu)} - “How fluent, grammatical, well formed and easy to understand are the generated utterances?”
    \item \textit{Meaning (Mea)} - “Are the input and the output referring or meaning the same thing?"
    \item \textit{Creativity (Cre)} - “How creative are the generated utterances?”
    \item \textit{Metaphoricity (Met)} - “How metaphoric are the generated utterances?” 
\end{enumerate}
For each one of these dimensions, a score ranging from 1 (very low) to 5 (very high) had to be assigned based on the evaluator's personal judgement. Example sentences that were taken from~\cite{MERMAID} were provided to the annotators, in order to clarify the assignment further. Two are shown below:
\begin{enumerate}
    \item \textit{The scream pierced the night}. Fluency: 4, Meaning: 5, Creativity: 4, Metaphoricity: 4;
    \item \textit{The wildfire swept through the forest at an amazing speed}. Fluency: 4, Meaning: 3, Creativity: 5, Metaphoricity: 4
\end{enumerate}

\begin{table}[ht]
  \centering
  \caption{Human evaluation for metaphorical sentences that were generated by our algorithm (top) or by humans (low). Scores are averaged across the sentences regarding fluency (Fl), meaning (Mn), creativity (Cr) and metaphoricity (Mt).}
    \begin{tabular}{llcccc}
     & & Fl & Mn & Cr & Mt\\
    \hline
    \multirow{2}{*}{A1}
    &Ours &\textbf{4.00} &3.65 &\textbf{3.11} &\textbf{3.41}  \\
    &Hum. &3.89 &\textbf{4.27} & 2.82 &3.21 \\
    \multirow{2}{*}{A2}
    &Ours & 4.39 &4.25 &\textbf{3.18} &\textbf{3.06}  \\
    &Hum. & \textbf{4.69} &\textbf{4.41} & 2.69 &2.97 \\
    \hline
    \multirow{2}{*}{Avg}
    &Ours & 4.20 &3.95 &\textbf{3.15} &\textbf{3.24}  \\
    &Hum. & \textbf{4.29} &\textbf{4.34} & 2.76 &3.09 \\
  \end{tabular}
  \label{tab:human}
\end{table}

Table~\ref{tab:human} shows the human evaluation for the system- and human- generated metaphorical sentences, regarding fluency, meaning, creativity and metaphoricity. The scores, averaged per creation source and dimension for each annotator, show that the first evaluator (A1) finds that the system-generated metaphors are better in three out of four dimensions. The second evaluator (A2) scored the human-generated metaphors higher in terms of fluency, but also scored the system-generated ones higher in terms of creativity and metaphoricity. The macro-averaged scores across the two annotators in the last two rows reflect this finding, showing that our system-generated metaphors are better in creativity and metaphoricity but lack in meaning preservation.

\begin{table*}[]
    \centering
     \caption{The three highest-scored human (H) and system (S) -generated metaphors. The latter outperform human-generated ones on average. We show the scores in a 1-5 scale, with 1 denoting the worst and 5 the best, that were assigned to each sentence for Fluency (Fl), Meaning (Mn), Creativity (Cr) and Metaphoricity (Mt). The tokens highlighted in bold are the words that are supposedly being used in a figurative way inside the sentences.}
    %\begin{adjustbox}{width=1.15\textwidth,center=\textwidth}
    \begin{tabular}{c|p{7cm}||c|c|c|c|c}
        & \textbf{Metaphorical sentence (metaphor in bold)} & \textbf{Original literal word} & \textbf{Fl} & \textbf{Mn} & \textbf{Cr} & \textbf{Mt} \\ \hline
        \textsc{s} & Day by day his heart within him \textbf{grew} more saturated with love and longing & \textbf{was} & 5 & 5 & 5 & 5 \\ \hline
        \textsc{s} & Through the green lanes of the country, where the tangled barberry-bushes \textbf{fluttered} their tufts of crimson berries & \textbf{tangled} & 5 & 5 & 5 & 5 \\ \hline
        \textsc{s} & Love the wind among the branches, and the rain-shower and the snow-storm, and the \textbf{roaring} of great rivers & \textbf{rushing} &5 &5 &4 &5 \\ \hline
        \textsc{h} & Headlines \textbf{scream} of pollution and dwindling natural resources	 & -- &5 &5 &4 &5 \\ \hline
        \textsc{h} & Musical creativity really \textbf{flowed} inside that family & -- &5 &5 &4 &4 \\ \hline
        \textsc{h} & This one scandal could very well \textbf{sink} his candidacy	& -- &5 &5 &4 &4 \\
    \end{tabular}
%    \end{adjustbox}
\label{tab:annotation}
\end{table*}

\section{Discussion}
\textbf{A quality assessment} is presented in Table~\ref{tab:annotation}, which shows three system-generated sentences that obtained the highest-score (ranking based on A1) and the respective three highest-scored human-generated ones, along with their four assigned scores. Although all the six sentences, human and system -generated, got an excellent score in fluency and meaning, our algorithm creates better metaphors with regards to creativity and metaphoricity. Two system-generated sentences out of three got an excellent creativity score with the third one obtaining a score equal to 4, while all human-generated sentences got a creativity score of 4.
All three system-generated sentences got a metaphoricity score of 5, while only one of the top human-generated sentences reached this score.\footnote{The similarity between the initial literal and the new metaphorical sentences that are constructed was computed with BERTScore~\cite{Zhang} and was found to be very high (0.99) for all topics, probably due to the fact that only a single word had to change per sentence.}

We believe that creativity is a very important dimension, which can facilitate human tasks, e.g., by providing inspiration. This is why it is worth noting that our approach not only shows promising results based on human evaluation, but also generates more creative metaphoric sentences than its human competitor.

The standard error of mean per dimension (Fl, Me, Cr, Mt) is respectively 0.09, 0.11, 0.16, 0.15 for the first annotator and 0.06, 0.10, 0.08, 0.08 for the second. The ones based on our system were only slightly higher. The mean absolute error between the annotations of the two annotators, per dimension, is respectively 1.04, 0.92, 1.35 and 1.24, reflecting the subjectiveness of the task: in fact, these differences are not statistically significant.

As far as the qualitative analysis of the results is concerned, the following are a few examples showing where the pipeline failed. In particular, we report examples from the Wikipedia music, Wikipedia technology and Gutenberg poetry corpus reconstructed sentences, whose newly generated verbs were mistakenly still identified as literal by our system in the final process' step. In fact, it is clear that these reconstructed words were characterized by a more metaphorical meaning compared to their original counterparts (thus, accomplishing the algorithm~\ref{alg:mdg}'s purpose).

Wikipedia music:
\begin{enumerate}
    \item "Music \textbf{drawn} solely from electronic generators was first produced in Germany in 1953" - \textit{Reconstructed sentence}: \textbf{drawn} was still classified as literal
    \item "Music \textbf{produced} solely from electronic generators was first produced in Germany in 1953" - \textit{Original literal sentence}
\end{enumerate}
Wikipedia technology - Example I:
\begin{enumerate}
    \item "The prehistoric invention of shaped stone tools \textbf{inspired} by the discovery of how to control fire increased sources of food..." - \textit{Reconstructed sentence}: \textbf{inspired} was still classified as literal
    \item "The prehistoric invention of shaped stone tools \textbf{followed} by the discovery of how to control fire increased sources of food..." - \textit{Original literal sentence}
\end{enumerate}
Wikipedia technology - Example II:
\begin{enumerate}
    \item "The invention of the wheel \textbf{encouraged} humans to travel in and control their environment" - \textit{Reconstructed sentence}: \textbf{encouraged} was still classified as literal
    \item "The invention of the wheel \textbf{helped} humans to travel in and control their environment" - \textit{Original literal sentence}
\end{enumerate}
Gutenberg poetry corpus:
\begin{enumerate}
    \item "Finally, my mother used to \textbf{rock} me to sleep..." - \textit{Reconstructed sentence}: \textbf{rock} was still classified as literal
    \item ""Finally, my mother used to \textbf{put} me to sleep..." - \textit{Original literal sentence}
\end{enumerate}

\subsection*{Improving metaphorical text classification}
Motivated by the promising human evaluation of our system-generated metaphors, we hypothesise that they might be beneficial as training material for metaphor-related tasks. To experiment with this hypothesis, a random sample of our new system-generated metaphorical sentences has been attached to the TroFi-X training set that we used to train the metaphorical sentence XLM-R classifier.\footnote{We employed TroFi-X for this experiment, since this dataset comprises nouns, verbs and adjectives, similarly to the new artificial data.} We also attached the same number of randomly sampled literal sentences, leading to 428 more training sentences in total (an increase of 37\%). Both the artificial metaphorical sentences and the literal ones have been extracted from Wikipedia and the Gutenberg Poetry Corpus. By fine-tuning the XLM-R metaphorical/literal sentence classifier on the increased training set, a percentage increase of all four classification metrics has been registered across TroFi-X over the respective scores of Table~\ref{tab:classification}: 3\% up in F1 (96.12\%), Precision (96.88\%) and Recall (95.38\%); 2.8\% in Accuracy (96.55\%). Simpler augmentation strategies, such as random instance duplication, yielded no improvement.\footnote{We used the same number of added instances and the results showed that the same number of true positives is achieved for literal texts but less for metaphorical ones.}

\subsection*{Metaphorical token recognition}
BERT and XLM-R can be used to successfully classify metaphorical sentences (Table~\ref{tab:classification}) and to reconstruct a metaphor through Masked Metaphor Modeling (MMM), with XLM-R achieving even the best reconstruction accuracy in one case (see T1 of TroFi-X in Table~\ref{tab:reconstruction}). Although reconstruction is based on the fact that the information of the location of the metaphor is already known (Section~\ref{ssec:methods}), we also assessed the ability of the BERT and XLM-R metaphorical sentence classifiers to recognize the exact location of the metaphor. Automated metaphor recognition could potentially allow the use of a dataset $I^{M|L}$ that will only comprise text level annotations without any token-level tags, such as the much larger TroFi dataset (Table~\ref{tab:dataset}).

We filtered the metaphorical sentences that were correctly classified (true positives) respectively by the fine-tuned BERT and XLM-R sentence classifiers and then we used the attention of the CLS token, in order to detect the location of the metaphor. In this study, we employed the fifth attention layer and the second to last (eleventh) head, since this combination yielded the best results in preliminary experiments, but we note that there are 144 possible layer-head combinations that could have also been investigated~\cite{Clark2019,Voita2019,Rogers2020}. The location of the metaphor, then, is simply considered to be the token of the sentence that received the maximum attention.
Table~\ref{tab:location_detection} provides the accuracy for this metaphor location detection task, which is the fraction of metaphorical sentences whose metaphor location was correctly detected. XLM-R is consistently better than BERT, while both models perform best in MOH-X and worse in TroFi. 

\begin{table}[pt]
    \centering
    \caption{Accuracy of BERT and XLM-R for metaphor location detection across the datasets}
    \begin{tabular}{ccccccc}
         & \multicolumn{1}{c}{\sc moh-x} & \multicolumn{1}{c}{\sc trofi} & \multicolumn{1}{c}{\sc trofi-x} \\\hline
         
         \sc bert &70.97&47.62&56.67\\
         \sc xlm-r &\bf 77.42&\bf 57.41&\bf 63.33\\
         
    \end{tabular}
    \label{tab:location_detection}
\end{table}

Three example MOH-X sentences are shown below with metaphorical tokens in bold and italics, and with XLM-R's attention heatmap in gray shade. In the first sentence, most of the attention was on the gold metaphorical verb. In the second one, it was on a part of the gold verb while in the third one it was on the gold noun (`soup') and the (not gold) adjective on the left (`hot').

\begin{enumerate}
    \item \textit{\textbf{He} \textbf{\colorbox[rgb]{.8,.8,.8}{marched}}} into the classroom and announced the exam.
    \item I \textit{\textbf{\colorbox[rgb]{.9,.9,.9}{wrest}led}} with this \textit{\textbf{decision}} for years.
    \item A \textit{\colorbox[rgb]{.85,.85,.85}{hot }\colorbox[rgb]{.95,.95,.95}{\textbf{soup}}} will \textit{\textbf{revive}} me.
\end{enumerate}

We note that only metaphorical words were ablated here, rather than any word as in the MLM's objective. We consider this use of Masked Language Modeling novel due to this explicit focus, which would have made word control and selection far more challenging.
The usefulness of metaphorical location recognition is indicated by the high accuracy (see Table~\ref{tab:location_detection}; it reaches up to 77.42\%), as it can unlock the development of larger training datasets in future work.

\subsection*{Ethical Considerations}
The following ethical considerations are worth our attention and perhaps further investigation in future work:
\begin{itemize}
    \item The proposed approach could in principle be used to create toxic (e.g., sarcastic or heavily ironic) text alternations about people. To avoid such misuse, the employed models were fine-tuned on data obtained from Wikipedia, which do not comprise abusive language or ironic speech.
    \item By implementing the proposed algorithm on data collected from the wild, one could end up sharing explicit details leaking information. One of the authors undertook a manual investigation of the data to verify that this was not the case in this study. Furthermore, we suggest that future use of the algorithm shall validate that the data will not contain sensitive user information that is not shared with the world. Examples of such information comprise health or negative financial status, racial or ethnic origin, religious or philosophical affiliation or beliefs, sexual orientation, trade union membership, alleged or actual commission of crime.
    \item In this study we opted for language models that are trained on data collected from the Web. These models may carry bias and have issues with abusive language~\cite{Sheng,Wallace}. We expect that the inductive bias of our models will limit inadvertent negative impacts. For example, BART is a conditional language model, which provides more control of the generated output. In any case, however, updating Algorithm~\ref{alg:mdg} with the addition of a text toxicity classifier \cite{Kiritchenko2021} could limit any unwanted outcomes.
    %The proposed approach can help with the generation of metaphorical text, providing resources, for example, to creative writing practitioners. 
    \item We note that ethical issues that go beyond the authors' knowledge may exist. Hence, and in order to allow future studies of possible weaknesses and vulnerabilities, along with the artificial data we release the source code and released trained models\footnote{The URL remains \url{hidden} to preserve the anonymity of the authors during the review process. The synthetic data and other material are now shared as supplementary material.}~\ref{appendix:shared_data}.%https://bit.ly/3I7Db76
\end{itemize}

\section{Conclusion}
This study presented an algorithm for transferring literal to metaphorical language, by employing metaphorical sentence classification and metaphor reconstruction. The obtained results showed that 24\%, 31\% and 56\% of the originally literal sentences get classified as metaphorical after masking and then reconstructing a noun, an adjective or a verb, respectively. 
Human evaluation on a mixed test set of system- and human- generated metaphorical sentences showed that we are able to generate synthetic metaphors that are considered on average as more creative and metaphorical than ones created by a human competitor. By using our synthetic metaphors to augment a metaphorical sentence classification dataset, we registered an F1 improvement of 3\% for an XLM-R metaphorical sentence classification baseline that was fine-tuned on the augmented dataset. The potential benefit of using a larger-scale version of our synthetic dataset, in order to improve metaphorical sentence classification further, will be studied in future work.
External reviewers have provided us with very useful feedback, which has improved this paper. Our extensive experiments concern a challenging natural language generation task that deserves more work, which we hope to facilitate with our shared resources. Finally, we consider that our intuitive suggested approach can inspire other researchers in this field and beyond, as in text simplification.

\subsection*{Limitations}
\begin{itemize}
    \item The proposed approach receives (likely) literal texts, that are ablated (masking) and then reconstructed as metaphorical ones. Human evaluation showed that the system-generated metaphors follow behind human-generated ones when assessing ``meaning''. However, we did not investigate the reasons behind this assessment, which could lead to suggestions for future improvements.
    \item We used artificial data, generated by employing our algorithm, in order to improve metaphorical text classification with data augmentation. The benefit was more clearly shown compared to simple oversampling techniques. We note, however, that several augmentation strategies could have been tested, such as using large pre-trained language models to generate text \cite{Bayer2021}. Such a study could reveal the potential of the proposed approach as an augmentation strategy.
    \item Extrinsic evaluation of this approach, which could improve downstream tasks, has not been explored in detail yet and will be studied in future work.
    \item Although label leaking could happen in our experiments, we note that this is a valid concern with any augmentation application. Nevertheless, all our data and code will be shared to allow further analysis and validation.
\end{itemize}
%\section*{Ethics Statement}
%\section*{Acknowledgements}

\bibliographystyle{unsrtnat}
\bibliography{references}

\appendix
\section{Appendix}
\label{appendix:shared_data}
\subsection{Shared Software and Data}
% the \\ insures the section title is centered below the phrase: AppendixA
All the exploited models, software and Python code have been shared and run using Google Colaboratory's GPUs.
The characteristics and the hyperparameter configurations for the best-performing model, the fine-tuned XLM-R metaphor classifier, are the following:
\begin{itemize}
    \item Model: fine-tuned XLMRobertaForSequenceClassification model from Transformers
    \item Tokenizer: XLMRobertaTokenizer from Transformers
    \item batch size = 32
    \item number of labels = 2
    \item optimizer = AdamW
    \item learning rate = 2e-5 (default value = 5e-5)
    \item adam\_epsilon = 1e-8 (default value)
    \item epochs = 10
    \item seed value = 42
\end{itemize}
Additional details such as the models' runtime (training, inference, etc.), validation performances and number of training and evaluation runs, depend on the datasets being used and can be found in comments inside the shared code for all experiments/approaches.

% the \\ insures the section title is centered below the phrase: Appendix B

All the datasets that were used and/or obtained during the experiments have been shared. The datasets are in English language and in .csv, .txt or .xlsx formats. Sentences from MOH-X, TroFi and TroFi-X are either labelled as metaphorical (label = 1) or literal (label = 0), and the datasets' structure is the following:
\begin{itemize}
    \item MOH-X: arg1, arg2, verb, sentence, verb\_idx, label
    \item TroFi: verb, sentence, verb\_idx, label
    \item TroFi-X: arg1, arg2, verb, sentence, verb\_stem, label
\end{itemize}
Each original and custom dataset's number of sentences can be found in the paper, along with an explanation of any data that were excluded, and all pre-processing steps.
Data initially scraped from Wikipedia and Gutenberg poetry corpus (then processed with masked metaphor modeling), as well as data obtained through augmentation techniques described in the paper, were kept in the following format:
\begin{itemize}
    \item sentence: scraped sentences with original or masked tokens
    \item label: 1 = metaphorical or 0 = literal, for the metaphor classifiers
\end{itemize}
Additional details such as those regarding train/validation/test splits ratios can be found in comments inside the shared code for all experiments/approaches.

\end{document}